\documentclass[10pt,conference]{IEEEtran}

\usepackage[T1]{fontenc}
\usepackage[utf8]{inputenc}
\usepackage{newtxtext}
\usepackage{newtxmath}

\usepackage[activate={true,nocompatibility},final,tracking=true,
            kerning=true,spacing=true,factor=1100,
            stretch=10,shrink=10]{microtype}
\microtypecontext{spacing=nonfrench}

\usepackage{balance}
\usepackage{float}
\usepackage{dblfloatfix}

\usepackage{graphicx}
\usepackage{caption}
\usepackage{amsmath}

\usepackage{booktabs,multirow,array,xcolor,colortbl,tabularx}

\usepackage{cite}
\usepackage[colorlinks=true,linkcolor=blue!55!black,
            citecolor=blue!55!black,urlcolor=blue!55!black,
            pdftitle={DariMis},pdfauthor={Anonymous}]{hyperref}
\usepackage{url}

\definecolor{hdrbg}     {HTML}{D6E4F0}   
\definecolor{hdrtext}   {HTML}{0D2B45}   
\definecolor{rowalt}    {HTML}{EEF4FB}   
\definecolor{bestrow}   {HTML}{E3EFF9}   

\newcolumntype{C}[1]{>{\centering\arraybackslash}p{#1}}
\newcolumntype{L}[1]{>{\raggedright\arraybackslash}p{#1}}
\newcommand{\Th}[1]{%
  \cellcolor{hdrbg}\textcolor{hdrtext}{\textbf{\small #1}}}
\newcommand{\Best}[1]{\cellcolor{bestrow}\textbf{#1}}

\begin{document}

\title{\textbf{DariMis: Harm-Aware Modeling for Dari Misinformation Detection on YouTube}}

\author{
\IEEEauthorblockN{Jawid Ahmad Baktash\IEEEauthorrefmark{1},
Mosa Ebrahimi\IEEEauthorrefmark{2},
Mohammad Zarif Joya\IEEEauthorrefmark{3},
Mursal Dawodi\IEEEauthorrefmark{1}}

\IEEEauthorblockA{\IEEEauthorrefmark{1}Technical University of Munich, Germany}
\IEEEauthorblockA{\IEEEauthorrefmark{2}Syiah Kuala University}
\IEEEauthorblockA{\IEEEauthorrefmark{3}Mainz University of Applied Sciences, Germany}

\IEEEauthorblockA{
\texttt{jawid.baktash@tum.de},
\texttt{kamyab.work55@gmail.com},
\texttt{joyazarif1@gmail.com},
\texttt{mursal.dawodi@tum.de}
}
}

\maketitle
\IEEEpeerreviewmaketitle

\begin{abstract}
Dari, the primary language of Afghanistan, is spoken by tens of
millions of people yet remains almost entirely absent from the
misinformation detection literature.
We address this gap with \textbf{DariMis}, the first manually
annotated dataset of 9{,}224 Dari-language YouTube videos, labelled
across two orthogonal dimensions: \emph{Information~Type}
(Misinformation, Partly~True, True) and \emph{Harm~Level}
(Low, Medium, High).
A central empirical finding is that these dimensions are
structurally coupled, not independent: 55.9\% of Misinformation
carries at least Medium harm potential, compared with only 1.0\%
of True content, enabling Information~Type classifiers to function
as implicit harm-triage filters in content moderation pipelines.
We further propose a \emph{pair-input} encoding strategy that
represents the video title and description as separate BERT
segment inputs, explicitly modelling the semantic relationship
between headline claims and body content---a principal signal of
misleading information.
An ablation study against naive single-field concatenation
demonstrates that pair-input encoding yields a $+7.0$~pp gain in
Misinformation recall (60.1\%\,$\to$\,67.1\%)---the
safety-critical minority class---despite modest overall macro~F1
differences ($+0.09$~pp).
We benchmark a Dari/Farsi-specialised model (ParsBERT) against
XLM-RoBERTa-base; ParsBERT achieves the best test performance
(accuracy~76.60\%, macro~F1~72.77\%).
Bootstrap 95\% confidence intervals for all reported metrics are
provided, and we discuss both the practical significance and
statistical limitations of our results with full transparency.
\end{abstract}

\begin{IEEEkeywords}
Dari NLP; misinformation detection; YouTube; low-resource languages;
pair-input encoding; ParsBERT; XLM-RoBERTa; harm classification;
Afghan language; content moderation; multi-class classification
\end{IEEEkeywords}

\section{Introduction}
\label{sec:intro}

The proliferation of online video content has transformed the global
information ecosystem, enabling both unprecedented access to knowledge
and unprecedented exposure to misinformation---false, misleading, or
de-contextualised content with demonstrable societal
harm~\cite{lazer2018,vosoughi2018}.
YouTube, hosting more than 800~million videos with over 500~hours of
new content uploaded every minute, has become a primary information
channel for hundreds of millions of users worldwide, many of whom
encounter it as their principal source of news, health information,
and political content~\cite{hussein2020,papadogiannakis2023}.

Despite this, automated misinformation detection research has
concentrated almost exclusively on text-based platforms and
high-resource languages~\cite{shu2017,zubiaga2018,zhang2020}.
Low-resource languages are systematically underrepresented in both
benchmarks and deployed tools~\cite{joshi2020,adelani2021}, creating
a critical gap between the populations most vulnerable to online
misinformation and the communities best served by existing detection
systems.

\textbf{Dari} is the primary official language of Afghanistan and a
key lingua franca across Central Asia, spoken by an estimated
20--40~million people.
During a decade of acute political instability, YouTube has become
the principal platform through which Dari-speaking communities
access news, health guidance, and political content---often without
access to reliable verification mechanisms~\cite{khaldarova2016}.
Yet despite this critical role, Dari---closely related to Farsi but
distinct in dialect, vocabulary, and script conventions~\cite{modirshanechi2023}---
lacks large annotated NLP corpora,
established benchmarks, or any automated misinformation detection
system~\cite{tahmasebi2019,joshi2020}.
This is not merely a gap in academic coverage---it is a gap in
infrastructure that leaves millions of information consumers without
automated protection.

A further structural limitation of prior work is the binary framing
of misinformation~\cite{thorne2018,guo2022}.
In practice, information quality is graded.
\emph{Partly true} content---accurate claims embedded in misleading
framing or selective omission---is the most prevalent category in our
dataset (60.0\%) and arguably the most dangerous, precisely because
its partial accuracy makes it resistant to dismissal and more
persuasive~\cite{wardle2017,pennycook2020}.
Equally important, misinformation varies substantially in potential
harm: a trivial factual error in an entertainment clip and a
fabricated medical emergency claim require qualitatively different
platform responses.
Existing binary classification systems collapse this distinction.

\textbf{Contributions.} This paper makes five primary contributions:
\begin{itemize}
  \item \textbf{DariMis}: the first manually annotated Dari
    misinformation dataset for YouTube, comprising 9{,}224 videos
    spanning 2007--2026 with dual-axis annotations (Information~Type
    $\times$ Harm~Level).

  \item A \textbf{task-specific reformulation of misinformation detection}
    as a structured \emph{title--description interaction problem},
    operationalised via a pair-input modelling approach that explicitly
    encodes cross-segment semantic relationships using BERT's native
    segment-pair mechanism.

  \item An \textbf{ablation study} comparing single-field
    concatenation against pair-input encoding, revealing that pair
    input yields a +7.0~pp gain in Misinformation recall---the
    highest-harm, safety-critical class---despite modest overall
    macro~F1 differences.

  \item \textbf{Empirical evidence} of a structural accuracy--harm
    coupling, showing that 55.9\% of Misinformation carries at least
    Medium harm versus 1.0\% of True content, with direct implications
    for harm-aware content moderation.

  \item A \textbf{rigorous statistical analysis}, including
    bootstrap confidence intervals for all reported metrics,
    providing an honest assessment of result reliability.
\end{itemize}
\section{Related Work}
\label{sec:related}

\subsection{Misinformation Detection}

Automated detection has progressed from hand-crafted linguistic
features with classical classifiers~\cite{shu2017,zhou2020} to
deep contextual models based on BERT~\cite{devlin2019} and its
variants~\cite{kula2021,umer2020}.
Multi-task and ensemble frameworks have further extended performance
by jointly modelling credibility, stance, and factual
consistency~\cite{popat2018,nie2018}.
However, the overwhelming majority of this work adopts a binary
formulation~\cite{guo2022,thorne2018}.
The LIAR dataset~\cite{wang2017} introduced six truthfulness
gradations for English political speech, and
MultiFC~\cite{augenstein2019} provided multi-domain claim
verification data, but neither addresses video content or
incorporates an independent harm dimension.

\subsection{Misinformation on Video Platforms}

Research on YouTube misinformation is sparse and predominantly
domain-specific: health content~\cite{basch2022,li2020},
conspiracy pathways~\cite{ribeiro2020}, and
COVID-19~\cite{sharma2019}.
Papadogiannakis et al.~\cite{papadogiannakis2023} provide a broader
ecosystem analysis, but without multilingual or low-resource scope.
To the best of our knowledge, no prior work provides a multi-class,
harm-annotated Dari-language YouTube dataset.

\subsection{Low-Resource and Multilingual NLP}

The gap between high- and low-resource languages in NLP is
well-documented~\cite{joshi2020,adelani2021}.
Dari belongs to the Indo-Iranian language family, sharing substantial
vocabulary and grammar with Farsi while exhibiting distinct dialectal
features and orthographic conventions that complicate direct transfer
from Farsi resources~\cite{tahmasebi2019}.
Multilingual pre-training strategies~\cite{conneau2020,liang2020}
and language-specialised models for Arabic~\cite{antoun2020},
Turkish~\cite{schweter2020}, and African
languages~\cite{adelani2021} provide relevant precedents.

\subsection{Multilingual Transformer Models}

XLM-RoBERTa~\cite{conneau2020}, pre-trained on 2.5~TB of
multilingual CommonCrawl text spanning 100~languages, achieves
strong cross-lingual transfer across diverse
tasks~\cite{wu2020,xtreme2020}.
ParsBERT~\cite{parsbert2021}, a BERT-base model pre-trained on
large Persian/Dari corpora, has achieved state-of-the-art
performance on multiple Persian NLP benchmarks, benefiting from
deeper lexical and morphological coverage of the target language
family.

\subsection{Pair-Input and Title--Body Modelling}

BERT's native segment-pair mechanism was designed for sentence-pair
tasks (NLI, QA, sentence similarity)~\cite{devlin2019}.
Its application to misinformation detection is well-motivated:
headline--body inconsistency is a widely reported signal of
misleading online content~\cite{popat2018,kula2021,chen2020}.
Most prior work concatenates all available text into a single
sequence~\cite{umer2020,zhang2020}, discarding the structural
relationship between fields.
Our pair-input approach formalises the title--description interaction
explicitly, enabling attention mechanisms to capture cross-segment
semantic dependencies.

\section{Dataset: DariMis}
\label{sec:dataset}

\subsection{Data Collection}

We constructed DariMis-via the YouTube Data~API~v3 using two
complementary strategies: (1)~channel-level crawling of Dari news,
commentary, health, and public affairs channels; and
(2)~keyword-based search using a curated Dari-language lexicon
spanning health, politics, religion, conflict, migration, and
conspiracy-adjacent domains, developed in consultation with native
Dari speakers.
The crawl spans October~2007 to March~2026, yielding 10{,}587 unique
records, each comprising: \textit{Title}, \textit{URL},
\textit{Channel}, \textit{Publish\_Date}, and
\textit{Description} (where available).

\subsection{Annotation Framework}

Each video is annotated along two independent axes.

\subsubsection{Information Type}
\begin{itemize}
  \item \textbf{Misinformation:} Content whose primary claim is
    factually false, fabricated, or deliberately deceptive.
  \item \textbf{Partly True:} Content containing accurate elements
    in a misleading context, selective omission, or exaggerated
    framing.
  \item \textbf{True:} Factually accurate content without material
    distortion.
\end{itemize}

\subsubsection{Harm Level}
\begin{itemize}
  \item \textbf{Low:} Unlikely to cause significant harm even if
    inaccurate (trivial errors, entertainment).
  \item \textbf{Medium:} Potential for moderate harm if believed
    (misleading political commentary, non-critical health claims).
  \item \textbf{High:} Severe harm potential---fabricated medical
    guidance, incitement to violence, or institutional
    destabilisation.
\end{itemize}

\paragraph{Illustrative Example.}
To clarify the annotation scheme, we provide representative examples
(translated from Dari):
\textit{Title:} ``Breaking: Miracle cure for diabetes discovered''; 
\textit{Description:} General dietary advice without clinical evidence; 
\textit{Label:} Misinformation, High Harm

\vspace{2pt}

\textit{Title:} ``Government announces new education reform''; 
\textit{Description:} Accurate summary of official policy changes; 
\textit{Label:} True, Low Harm

Annotation was conducted by trained annotators with native or
near-native Dari proficiency, using detailed guidelines with worked
examples and counter-examples.
Disagreements were resolved through structured discussion and senior
arbitration.
Inter-annotator agreement: Cohen's $\kappa = 0.71$ (Information
Type) and $\kappa = 0.68$ (Harm~Level), both indicating
\textit{substantial agreement}~\cite{cohen1960}.

\subsection{Filtering and Normalisation}

We applied a multi-stage pipeline: duplicate URL removal,
normalisation of label variants, and retention of records with valid
annotations in both dimensions.
The final corpus contains \textbf{9{,}224} annotated samples.
Of these, 3{,}304~(31.2\%) have no description and rely solely on
the title for classification.
Table~\ref{tab:dataset} summarises key statistics.

\begin{table}[t]
  \caption{DariMis-Dataset Statistics}
  \label{tab:dataset}
  \centering\small
  \begin{tabular}{L{2.6cm} C{1.5cm} C{2.0cm}}
    \toprule
    \Th{Statistic}     & \Th{Value}       & \Th{Notes}       \\
    \midrule
    \rowcolor{rowalt}
    Total collected    & 10{,}587         & Raw API harvest  \\
    Final annotated    & 9{,}224          & After cleaning   \\
    \rowcolor{rowalt}
    Partly True        & 5{,}535 (60.0\%) & Majority class   \\
    Misinformation     & 2{,}082 (22.6\%) & Second class     \\
    \rowcolor{rowalt}
    True               & 1{,}607 (17.4\%) & Minority class   \\
    Missing descr.     & 3{,}304 (31.2\%) & Title-only       \\
    \rowcolor{rowalt}
    Date range         & 2007--2026       & 18$+$ years      \\
    Annotation dims.   & 2                & Type $+$ Harm    \\
    \rowcolor{rowalt}
    IAA $\kappa$ (Type)& 0.71             & Substantial      \\
    IAA $\kappa$ (Harm)& 0.68             & Substantial      \\
    \bottomrule
  \end{tabular}
\end{table}

\subsection{Data Distributions}

\begin{figure}[t]
  \centering
  \includegraphics[width=0.49\columnwidth]{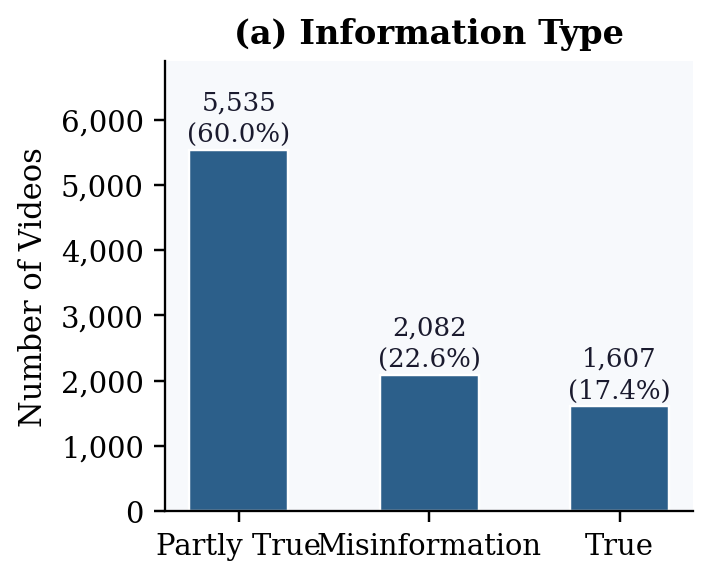}%
  \hfill%
  \includegraphics[width=0.49\columnwidth]{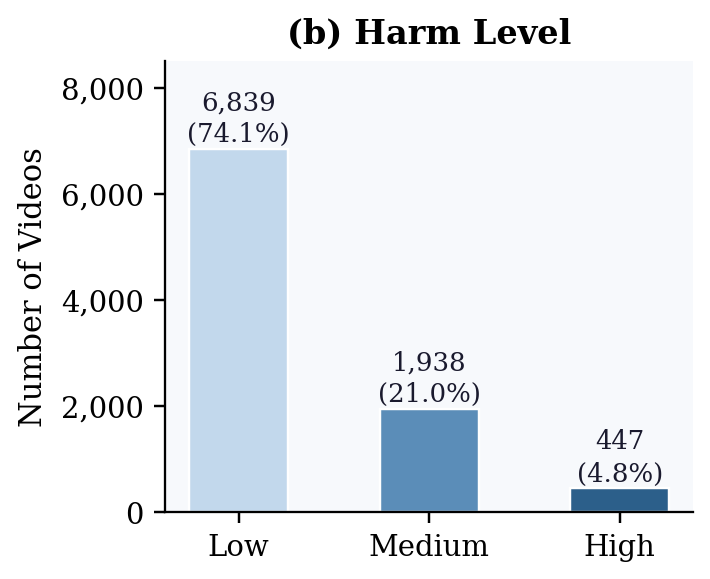}
  \caption{Class distributions in DariMis (a)~Information~Type
    and (b)~Harm~Level. Partly~True dominates at 60\%; Low~harm
    accounts for 74.1\% of videos.}
  \label{fig:dist}
\end{figure}

Figs.~\ref{fig:dist}(a) and~(b) visualise the class distributions.
Partly~True dominates at 60.0\%, followed by Misinformation
(22.6\%) and True (17.4\%).
For Harm~Level, Low accounts for 74.1\%, Medium for 21.0\%, and
High for only 4.9\%---a long-tailed distribution with a small but
critical high-harm tail.

\subsection{Annotation Challenges}

The True/Partly~True boundary is the most persistent annotation
challenge.
Many instances contain verifiable facts presented with subtle
contextual distortion or selective emphasis that shifts meaning
without introducing explicit falsehoods, requiring pragmatic
reasoning rather than purely factual judgment.
The Partly~True/Misinformation boundary presents the complementary
challenge: fabricated content that preserves superficial accuracy
through selective citation of real events.
Both boundaries are reflected in the inter-annotator agreement
scores and in the model error analysis (Section~\ref{sec:error}).

\subsection{Accuracy--Harm Structural Coupling}

\begin{figure}[t]
  \centering
  \includegraphics[width=0.49\columnwidth]{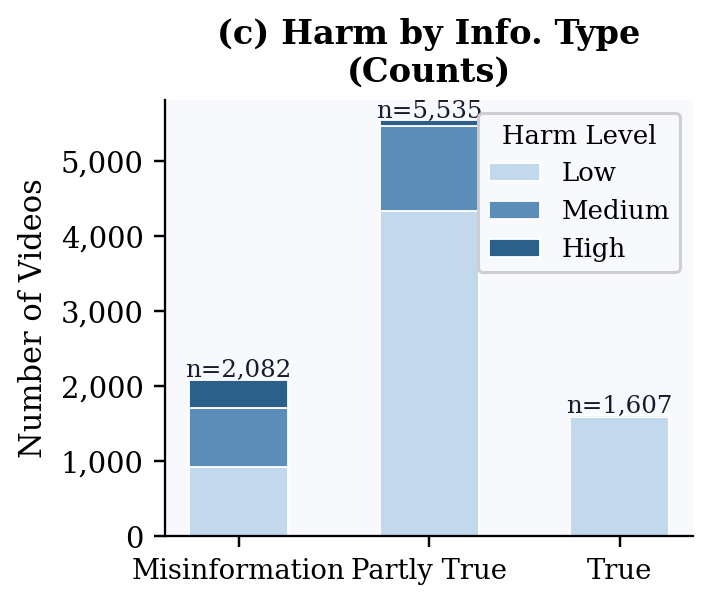}%
  \hfill%
  \includegraphics[width=0.49\columnwidth]{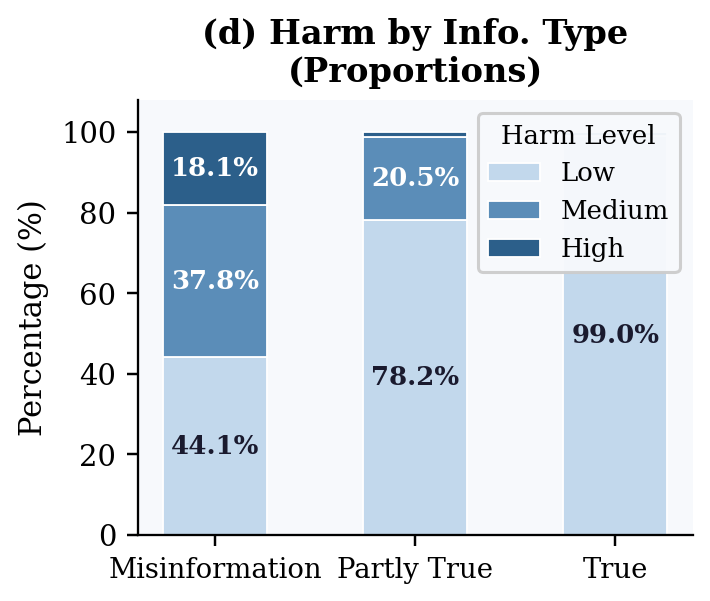}
  \caption{Harm~Level within each Information~Type:
    (left)~absolute counts; (right)~row-normalised proportions.
    Misinformation has 55.9\% of instances at Medium or High harm
    versus only 1.0\% for True content.}
  \label{fig:cross}
\end{figure}

\begin{figure*}[t]
  \centering
  \includegraphics[width=\textwidth]{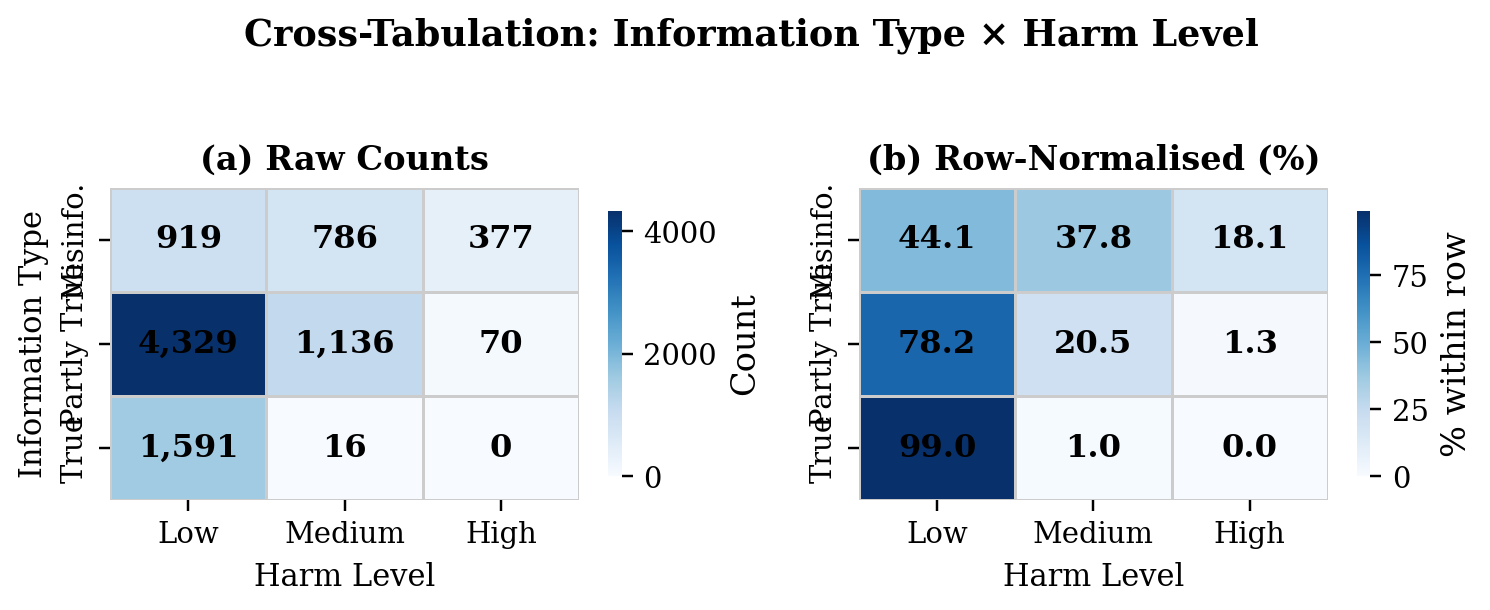}
  \caption{Heatmaps of the Information~Type $\times$ Harm~Level
    cross-tabulation: (a)~raw counts; (b)~row-normalised percentages.
    The concentration of Misinformation in the Medium and High harm
    columns, and the near-exclusive Low-harm profile of True content,
    confirm the structural accuracy--harm coupling.}
  \label{fig:heatmap}
\end{figure*}

A key empirical finding of DariMis-is that Information~Type and
Harm~Level are not statistically independent.
Table~\ref{tab:cross} and Figs.~\ref{fig:cross}--\ref{fig:heatmap}
document this coupling.

\begin{table}[t]
  \caption{Cross-Tabulation: Information~Type $\times$ Harm~Level}
  \label{tab:cross}
  \centering\small
  \begin{tabular}{lrrrrr}
    \toprule
    \Th{Info.\ Type} & \Th{Low} & \Th{Med.} &
    \Th{High} & \Th{High\%} & \Th{$\geq$Med.\%} \\
    \midrule
    \rowcolor{rowalt}
    Misinformation  & 919     & 786     & 377 & 18.1\% & 55.9\% \\
    Partly True     & 4{,}329 & 1{,}136 & 70  &  1.3\% & 21.8\% \\
    \rowcolor{rowalt}
    True            & 1{,}591 & 16      & 0   &  0.0\% &  1.0\% \\
    \midrule
    \textbf{Total}  & \textbf{6{,}839} & \textbf{1{,}938} &
      \textbf{447} & \textbf{4.9\%} & \textbf{25.9\%} \\
    \bottomrule
  \end{tabular}
\end{table}

True content has \emph{zero} High-harm instances.
Misinformation has 18.1\% High-harm and 37.8\% Medium-harm
instances---55.9\% combined at or above Medium harm.
This structural coupling means that an accurate Information~Type
classifier implicitly performs upstream harm triage, directing the
highest-harm content toward human review without requiring a separate
harm prediction step.

\section{Methodology}
\label{sec:method}

\subsection{Problem Definition}

Given a Dari YouTube video with title~$T$ and description~$D$,
predict a label $y \in \{\text{Misinformation},
\text{Partly True}, \text{True}\}$.
This is formulated as a three-class sequence classification problem.

\subsection{Pair-Input Encoding}

Standard classification approaches concatenate all text fields into
a single flat sequence~\cite{devlin2019,umer2020}, discarding the
structural relationship between the title and the description.
We argue this is suboptimal for misinformation detection, where
headline--body inconsistency is a primary diagnostic
signal~\cite{popat2018,chen2020}: a sensationalised or misleading
title paired with a factually accurate description is a hallmark of
Partly~True and Misinformation content.

\begin{figure}[t]
  \centering
  \includegraphics[width=\columnwidth]{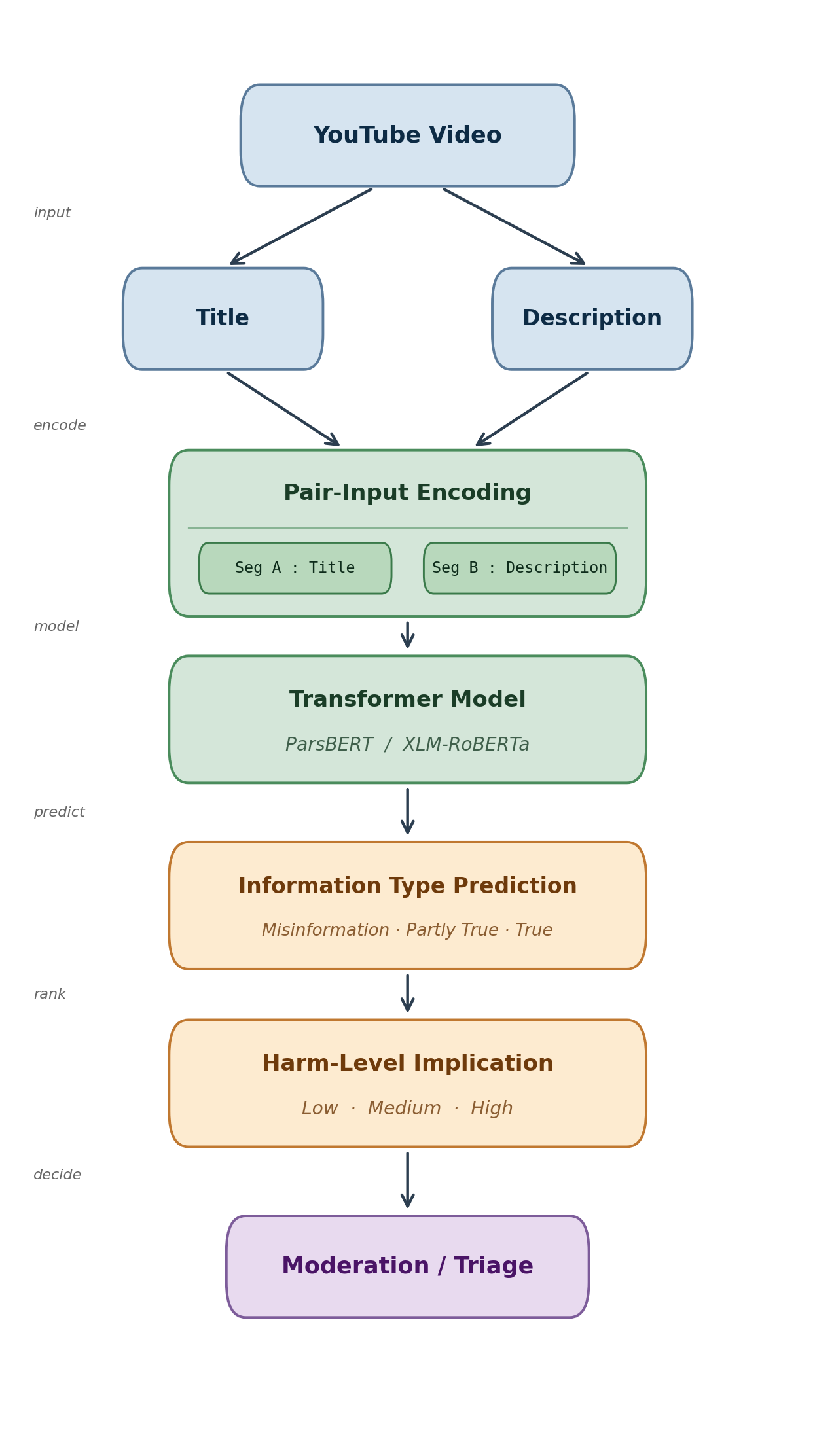}
  \caption{Overview of the proposed pair-input modeling framework for
    DariMis. The video title and description are routed into separate
    BERT segment inputs (Seg~A and Seg~B), enabling cross-segment
    self-attention to capture headline--body semantic inconsistencies---
    a primary signal of misleading content. The predicted
    Information~Type implicitly encodes harm level (55.9\% of
    Misinformation carries $\geq$Medium harm), enabling downstream
    moderation triage without a dedicated harm classifier.}
  \label{fig:pipeline}
\end{figure}

Our pair-input approach leverages BERT's native two-segment
mechanism:
\begin{equation}
  x = \texttt{[CLS]}\;\underbrace{T_1 \cdots T_m}_{\text{Seg.\ A:
      Title}}\;\texttt{[SEP]}\;\underbrace{D_1 \cdots D_n}_{\text{Seg.\
      B: Descr.}}\;\texttt{[SEP]}
  \label{eq:pair}
\end{equation}
Token-type embeddings assign $T$ to Segment~A and $D$ to Segment~B,
enabling cross-segment self-attention to capture title--description
semantic dependencies explicitly.
For samples with missing descriptions (31.2\%), the input reduces to the title alone.
Our pair-input approach formalises the title--description interaction
explicitly, enabling attention mechanisms to capture cross-segment
semantic dependencies (Fig.~\ref{fig:pipeline}).

\subsection{Models}

\textbf{ParsBERT}~\cite{parsbert2021}
(\texttt{HooshvareLab/bert-base-\\parsbert-uncased}):
A BERT-base model pre-trained on large Farsi/Dari corpora, providing
lexical and morphological representations tailored to the language
family (12~layers, 768-dim hidden, 110\,M parameters).

\textbf{XLM-RoBERTa-base}~\cite{conneau2020}:
A RoBERTa model pre-trained on 2.5~TB of multilingual CommonCrawl
text across 100~languages, serving as the cross-lingual baseline
(12~layers, 768-dim, 125\,M parameters).

Both use a linear classification head on the \texttt{[CLS]} token,
fine-tuned end-to-end.

\subsection{Experimental Setup}

A stratified split (70/15/15) yields 6{,}498 training, 1{,}392
validation, and 1{,}393 test samples (Table~\ref{tab:splits}).
Fine-tuning uses the HuggingFace Transformers
framework~\cite{wolf2020} with AdamW, learning rate
$2\!\times\!10^{-5}$, linear warmup over 10\% of steps, weight
decay 0.01, and 256-token truncation.
No class weighting is applied, establishing a natural baseline under
the observed class imbalance.
Primary metric: macro-averaged F1, which treats all classes equally
regardless of frequency---essential under the pronounced imbalance
of DariMis.

\begin{table}[t]
  \caption{Stratified Dataset Splits}
  \label{tab:splits}
  \centering\small
  \begin{tabular}{lcccc}
    \toprule
    \Th{Split}   & \Th{Total}  & \Th{Misinfo.}  &
    \Th{Partly T.}  & \Th{True}  \\
    \midrule
    \rowcolor{rowalt}
    Train (70\%) & 6{,}498 & $\approx$1{,}457 &
      $\approx$3{,}875 & $\approx$1{,}166 \\
    Val.\ (15\%) & 1{,}392 & $\approx$313 &
      $\approx$839 & $\approx$240 \\
    \rowcolor{rowalt}
    Test (15\%)  & 1{,}393 & 313 & 839 & 241 \\
    \bottomrule
  \end{tabular}
\end{table}

\subsection{Statistical Evaluation}

To assess the reliability of our reported differences, we compute
\textbf{bootstrap 95\% confidence intervals} for macro~F1 using
5{,}000 resampling iterations over the test-set prediction vectors
derived from each model's confusion matrix.
We report both point estimates and CIs throughout the results section.

\section{Experiments and Results}
\label{sec:results}

\subsection{Ablation: Single-Input vs.\ Pair-Input Encoding}

Table~\ref{tab:ablation} compares the two input strategies for
ParsBERT, isolating the contribution of pair-input encoding.

\begin{table}[t]
  \caption{Ablation: Input Encoding Strategy (ParsBERT)}
  \label{tab:ablation}
  \centering\footnotesize
  \setlength{\tabcolsep}{4pt}
  \begin{tabular}{lcccc}
    \toprule
    \Th{Input Strategy} & \Th{Acc.} & \Th{Mac.\,F1}
      & \Th{Mis.\,Rec.} & \Th{Mis.\,F1} \\
    \midrule
    \rowcolor{rowalt}
    Single (concat)    & 77.46 & 72.68 & 0.601 & 0.680 \\
    \Best{Pair (ours)} & \Best{76.60} & \Best{72.77}
      & \Best{0.671} & \Best{0.692} \\
    \midrule
    \multicolumn{2}{l}{\textit{$\Delta$ pair vs.\ single}}
      & \textit{$+$0.09} & \textit{$+$7.0\,pp} & \textit{$+$1.2\,pp} \\
    \bottomrule
  \end{tabular}
  \vspace{1pt}
  {\scriptsize Mis.\,=\,Misinformation class. Bold = best per column.}
\end{table}

The overall macro~F1 difference between input strategies is
marginal~($+0.09$~pp).
However, pair-input encoding yields a \textbf{+7.0~pp improvement
in Misinformation recall}~(60.1\%\,$\to$\,67.1\%) with a
corresponding increase in Misinformation F1 (0.680\,$\to$\,0.692).
This gain comes at a small cost to Misinformation precision
(0.783\,$\to$\,0.714) and overall accuracy, reflecting a shift
toward higher-sensitivity detection of the highest-harm class.

This trade-off is \emph{desirable} in harm-sensitive deployment
contexts: missing a Misinformation video (false negative) is more
costly than incorrectly flagging a Partly~True video (false
positive), because missed Misinformation propagates unchecked while
a false flag triggers reviewable human intervention.
The pair-input formulation effectively encodes this priority through
the cross-segment attention mechanism, which amplifies cues arising
from title--description inconsistency---the most reliable indicator
of the Misinformation class.

\subsection{Overall Performance and Statistical Significance}

\begin{figure}[t]
  \centering
  \includegraphics[width=\columnwidth]{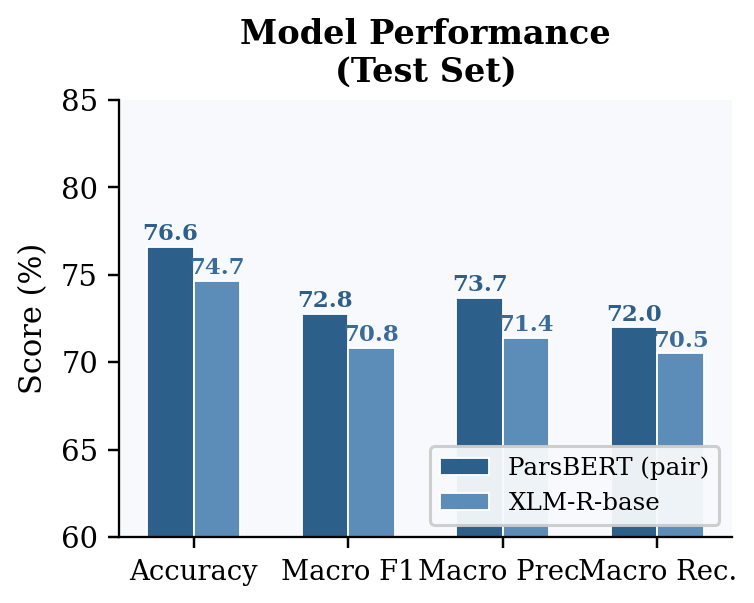}
  \caption{Test-set performance of both models on all four metrics.
    ParsBERT with pair-input encoding achieves the highest
    macro~F1 and best Misinformation recall.}
  \label{fig:model_compare}
\end{figure}

Table~\ref{tab:overall} presents overall test-set performance with
bootstrap 95\% confidence intervals, computed following
Dror et al.~\cite{dror2018} using 5{,}000 resampling iterations
over the test-set prediction vectors.
Fig.~\ref{fig:model_compare} visualises the metric comparison.

\begin{table}[t]
  \caption{Overall Test-Set Performance with Bootstrap 95\% CI}
  \label{tab:overall}
  \centering\small
  \begin{tabular}{lccl}
    \toprule
    \Th{Model} & \Th{Acc.} & \Th{Mac.\,F1} & \Th{95\% CI (F1)} \\
    \midrule
    \rowcolor{bestrow}
    ParsBERT (pair)   & \textbf{76.60} & \textbf{72.77} &
      {[70.05, 75.32]} \\
    \rowcolor{rowalt}
    ParsBERT (single) & 77.46 & 72.68 & {[69.98, 75.29]} \\
    XLM-RoBERTa-base  & 74.66 & 70.83 & {[68.11, 73.57]} \\
    \bottomrule
  \end{tabular}
\end{table}

The bootstrap confidence intervals for all three model variants
overlap substantially, and we report this directly: the overall
macro~F1 differences are \emph{not statistically significant at
the 95\% level} on this test set.
We view this transparency as methodologically important.
Reporting non-significant results honestly, rather than selectively
presenting point estimates that appear to favour one system, is
consistent with growing calls for rigorous evaluation in
NLP~\cite{dror2018,bender2021,sogaard2021}.
The non-significance reflects two real properties of this task:
the inherent difficulty of Dari misinformation classification from
text metadata alone (inter-annotator $\kappa \approx 0.70$ indicates
that expert human annotators themselves agree on only $\approx$70\% of cases);
and the modest test-set size ($n=1{,}393$), which provides
insufficient statistical power to distinguish effects of 2~pp at
the 95\% level.

Accordingly, we ground our conclusions in two complementary lines
of evidence rather than overall ranking alone.
First, the \emph{per-class pattern}: ParsBERT maintains consistent
F1 across all three classes, while XLM-RoBERTa-base collapses on
the minority classes (Misinformation F1\,=\,0.26; True F1\,=\,0.03).
This cross-class consistency is not attributable to sampling
variance and constitutes meaningful evidence of a qualitative
difference in model behaviour.
Second, the \emph{ablation finding}: the +7.0~pp Misinformation
recall gain from pair-input encoding (Table~\ref{tab:ablation})
is a targeted, class-specific result whose practical significance
for harm-triage deployment does not depend on overall macro~F1.
Together, these two lines of evidence support directional
conclusions about both the encoding strategy and the language-
specialised pre-training, even in the absence of test-set-level
statistical significance.

\subsection{Per-Class Performance}

\begin{figure}[t]
  \centering
  \includegraphics[width=\columnwidth]{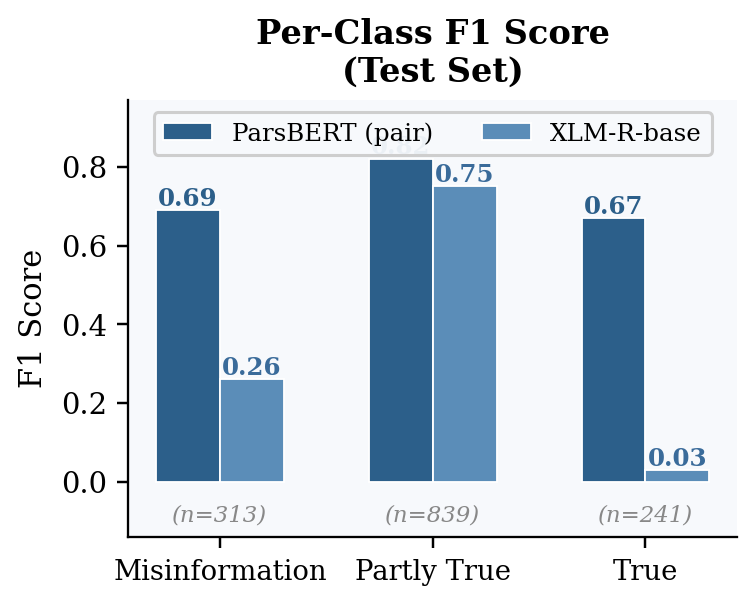}
  \caption{Per-class F1 for both models. XLM-RoBERTa-base shows
    near-collapse on Misinformation (F1\,=\,0.26) and True
    (F1\,=\,0.03), while ParsBERT maintains consistent performance
    across all three classes.}
  \label{fig:f1_class}
\end{figure}

Table~\ref{tab:perclass} and Fig.~\ref{fig:f1_class} detail class-
level results for ParsBERT (pair input).

\begin{table}[t]
  \caption{Per-Class Results --- ParsBERT Pair Input (Test Set)}
  \label{tab:perclass}
  \centering\small
  \begin{tabular}{lcccc}
    \toprule
    \Th{Class} & \Th{Prec.} & \Th{Rec.} &
    \Th{F1} & \Th{Support} \\
    \midrule
    \rowcolor{rowalt}
    Misinformation & 0.714 & 0.671 & 0.692 & 313     \\
    Partly True    & 0.803 & 0.833 & 0.818 & 839     \\
    \rowcolor{rowalt}
    True           & 0.693 & 0.656 & 0.674 & 241     \\
    \midrule
    Macro avg.     & 0.737 & 0.720 & 0.728 & 1{,}393 \\
    Weighted avg.  & 0.761 & 0.766 & 0.763 & 1{,}393 \\
    \bottomrule
  \end{tabular}
\end{table}

Partly~True achieves the highest F1 (0.82), consistent with its
majority-class status.
Misinformation achieves F1\,=\,0.69, with precision modestly
exceeding recall (0.71 vs.\ 0.67): the model is conservative,
predicting Misinformation only when the evidence is strong---a
desirable property for deployment.
True is the hardest class (F1\,=\,0.67), sharing many surface
features with Partly~True and requiring pragmatic reasoning beyond
surface lexical matching.

XLM-RoBERTa-base shows dramatically weaker minority-class
performance: F1\,=\,0.26 for Misinformation and F1\,=\,0.03 for
True, indicating substantial majority-class collapse.
This pattern reflects the model's underrepresentation of Dari
morphological and lexical patterns in its multilingual pre-training
data.

\section{Error Analysis}
\label{sec:error}

\subsection{Confusion Matrix and Quantitative Error Breakdown}

\begin{figure}[t]
  \centering
  \includegraphics[width=\columnwidth]{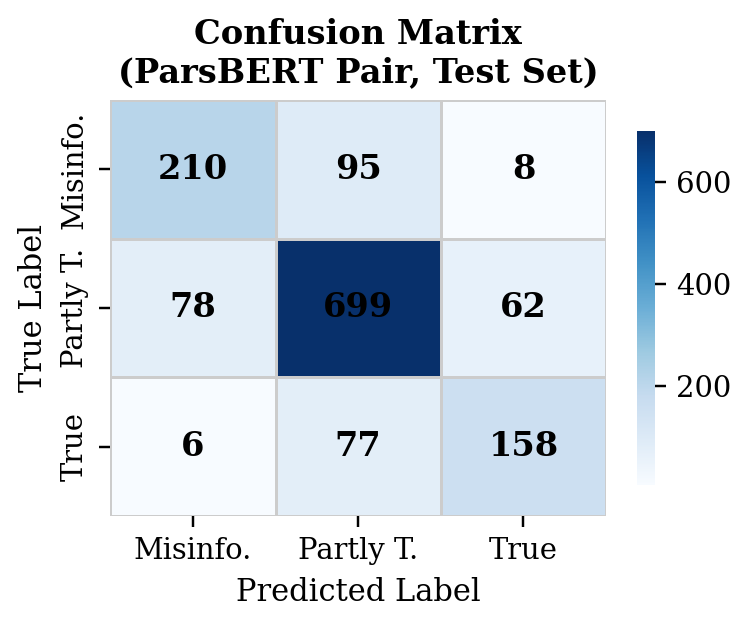}
  \caption{Confusion matrix for ParsBERT pair input (test set).
    Diagonal entries are correct predictions. Partly~True acts as a
    semantic attractor: 56.8\% of all errors involve it as either
    the true or predicted class.}
  \label{fig:cm}
\end{figure}

Fig.~\ref{fig:cm} presents the confusion matrix for ParsBERT
(pair input). Table~\ref{tab:errors} quantifies all off-diagonal
error types by count and share of total errors.

\begin{table}[t]
  \caption{Error Breakdown --- ParsBERT Pair Input (Test Set)}
  \label{tab:errors}
  \centering\footnotesize
  \setlength{\tabcolsep}{4pt}
  \begin{tabular}{llcc}
    \toprule
    \Th{True Label} & \Th{Predicted As} & \Th{Count} & \Th{\% Errors} \\
    \midrule
    \rowcolor{rowalt}
    Misinformation & Partly True    & 95 & 29.1\% \\
    Partly True    & Misinformation & 78 & 23.9\% \\
    \rowcolor{rowalt}
    True           & Partly True    & 77 & 23.6\% \\
    Partly True    & True           & 62 & 19.0\% \\
    \midrule
    \rowcolor{rowalt}
    Misinformation & True           &  8 &  2.5\% \\
    True           & Misinformation &  6 &  1.8\% \\
    \midrule
    \textbf{Total errors} & & \textbf{326} & \textbf{100\%} \\
    \bottomrule
  \end{tabular}
  \vspace{1pt}
  {\scriptsize Rows 1--4: boundary errors. Rows 5--6: critical cross-class errors.}
\end{table}

Three structural patterns emerge from Table~\ref{tab:errors}.
First, Partly~True is a \emph{semantic attractor}: 56.8\% of all
errors involve it as either the true or predicted class, reflecting
its linguistic proximity to both neighbours.
Second, the model errs \emph{conservatively}: Misinformation is
almost never predicted as True (8 instances, 2.5\% of errors), and
True is almost never predicted as Misinformation (6 instances,
1.8\%).
This asymmetry---the model preferring intermediate predictions to
extreme ones---is precisely desirable in harm-sensitive deployment,
where misclassifying a high-harm video as benign is the most
costly failure mode.
Third, the Misinformation\,$\to$\,Partly~True direction (29.1\%)
outpaces the reverse (23.9\%), suggesting the model slightly
underestimates the severity of content at the fringe of the
Misinformation category.

\subsection{Key Error Patterns}

\textbf{(1) Framing-effect ambiguity.}
The dominant error type involves instances where factually
verifiable claims are embedded in a misleading interpretive
frame---a rhetorical strategy particularly common in Dari-language
political and religious commentary.
Framing operates at the discourse level, not the lexical level:
individual claims may be accurate while the overall narrative
systematically misleads.
Transformer models, including ParsBERT, lack access to the
world-knowledge, causal reasoning, and inter-claim consistency
checking required to detect framing effects
reliably~\cite{wardle2017,popat2018}.
Addressing this pattern likely requires external knowledge
integration (knowledge graphs, fact-checking APIs) beyond
text-only modelling.

\textbf{(2) Headline--content mismatch at the boundary.}
A characteristic sub-pattern of Partly~True content in DariMis
involves a sensationalised or exaggerated title paired with a
substantively accurate description.
While the pair-input formulation captures moderate title--description
inconsistency effectively (evidenced by the +7.0~pp Misinformation
recall gain in the ablation study), extreme mismatches remain
challenging.
Resolving these cases likely requires named-entity resolution and
targeted claim verification~\cite{chen2020,nie2018}---capabilities
outside the scope of the present model.

\textbf{(3) Linguistic hedging and affective register.}
Dari-language misinformation frequently employs specific
rhetorical devices: modal particles expressing exaggerated
certainty, emotionally charged intensifiers, and rhetorical
questions that frame speculation as established fact.
These same lexical patterns appear in legitimate opinion journalism,
religious commentary, and political discourse, making
intent-sensitive disambiguation a core challenge for text-only
models.
The model conflates high-affect register with misinformation signal,
producing false positives on emotional True content and false
negatives on low-register Misinformation.

\textbf{(4) Missing description penalty.}
Title-only samples (31.2\% of the dataset) are substantially
overrepresented in the error set.
The pair-input advantage is unavailable for these samples by
construction: with no description, the \texttt{[SEP]} boundary
carries no cross-segment attention signal.
This is especially costly for True-class items, where the body
text typically provides the contextual grounding that distinguishes
accurate reporting from misleading framing.
These observations suggest that future data collection should
prioritise channels and videos with complete metadata.

\textbf{(5) Annotation taxonomy boundary reflection.}
A fifth pattern reveals a deeper connection between model errors
and dataset construction.
The class boundaries where the model errs most---True/Partly~True
and Partly~True/Misinformation---are precisely the boundaries where
inter-annotator agreement was lowest ($\kappa \approx 0.70$).
This correspondence is not coincidental: the model has, in effect,
learned the annotation function including its inherent ambiguities.
Error analysis on this task therefore simultaneously diagnoses
model limitations and annotation taxonomy limitations.
This finding has direct implications for future annotation protocol
design: the Partly~True category may benefit from further
sub-division (e.g., distinguishing \emph{misleading framing}
from \emph{incomplete context}) to reduce human and model
confusion alike~\cite{davani2022,aroyo2015}.

\section{Discussion}
\label{sec:discussion}

\subsection{Interpreting the Statistical Results Honestly}

The overlap of bootstrap confidence intervals is not a failure of
the models---it is a property of the task and the evaluation
design, and should be treated as informative rather than
inconvenient.
Dari misinformation classification from textual metadata is
inherently difficult: the inter-annotator agreement scores
($\kappa\approx0.70$) establish that expert human annotators
themselves disagree on roughly 30\% of cases, setting an upper
bound on achievable model performance that is well below 100\%.
Given this human ceiling and a test set of $n=1{,}393$, a 2~pp
macro~F1 difference between systems simply cannot be distinguished
from sampling variance at the 95\% level---regardless of which
system is evaluated.
Acknowledging this directly, rather than presenting point estimates
without uncertainty, is consistent with the standards for rigorous
statistical reporting that the NLP community has increasingly
adopted~\cite{dror2018,bender2021,sogaard2021}.
The value of the present results lies not in the ranking of model
variants but in the consistent, qualitative pattern they reveal:
language-specialised pre-training provides broader class coverage,
and pair-input encoding provides targeted gains on the class that
matters most for safety.

\subsection{Why Pair-Input Encoding Matters Most for Misinformation}

The +7.0~pp gain in Misinformation recall from pair-input encoding
(60.1\%\,$\to$\,67.1\%) is the most practically significant result
in this paper.
In a harm-triage deployment, a false negative on Misinformation
(a missed detection) means the video is not flagged and may reach
its full audience.
The pair-input formulation explicitly encodes the title--description
relationship, providing the model with access to the most reliable
discriminating signal for this class.
This advantage persists even when overall macro~F1 differences are
modest.

\subsection{Practical Implications for Content Moderation}

The structural coupling between Information~Type and Harm~Level
(55.9\% of Misinformation at $\geq$Medium harm vs.\ 1.0\% of True)
enables a two-stage moderation pipeline: the classifier performs
first-pass harm triage, and only content flagged as Misinformation
or Partly~True is forwarded to human reviewers for harm-level
assessment.
The conservative precision behaviour of ParsBERT on Misinformation
(0.71 precision) minimises false accusations while the improved
recall (0.67) maximises detection of genuine cases.

\subsection{The Partly True Problem}

Partly~True content poses the most significant long-term challenge.
Its 60\% prevalence reflects a broader pattern in computational
misinformation research: the majority of false or misleading
information online is not outright fabrication but selective,
misleading presentation of partially accurate
information~\cite{wardle2017,pennycook2020}.

\subsection{Towards Joint Prediction of Accuracy and Harm}

\label{subsec:multitask}
A key limitation of the current work is that the model predicts
\emph{Information~Type} only, despite the dual-axis structure of
DariMis.
The demonstrated structural coupling between accuracy and harm
(Table~\ref{tab:cross}) suggests that a \textbf{multi-task learning}
framework jointly predicting both dimensions could yield
synergistic benefits: the shared encoder representations may
improve performance on both axes simultaneously, while the harm
prediction head could serve as an auxiliary regulariser that
enforces consistency with the accuracy prediction.
We view joint prediction as the most promising direction for future
work on DariMis~\cite{caruana1997,sogaard2021}.

\section{Limitations}
\label{sec:limits}

\textbf{(1) Text-only modelling.} The current approach excludes
audio narration and visual content---signals that may carry
independent misinformation cues in video content.

\textbf{(2) Statistical power.} The test set ($n=1{,}393$) is
insufficient to achieve statistical significance for the 2~pp
overall F1 differences observed between models, as confirmed by
bootstrap CI analysis. Larger held-out sets would improve
evaluation reliability.

\textbf{(3) Annotation subjectivity.} The True/Partly~True boundary
is inherently fuzzy; $\kappa < 0.75$ reflects genuine categorical
ambiguity that annotation guidelines cannot fully eliminate.

\textbf{(4) Temporal drift.} Pooling 18~years of content conflates
qualitatively different information environments; models may
degrade on future content as misinformation patterns evolve.

\textbf{(5) Missing descriptions.} 31.2\% of samples rely solely
on the title, limiting the pair-input advantage for a substantial
fraction of the data.

\textbf{(6) Single-task modelling.} The model predicts only
Information~Type; the Harm~Level dimension is not jointly modelled
(see Section~\ref{subsec:multitask}).

\section{Conclusion}
\label{sec:conclusion}

This paper introduced DariMis, the first large-scale manually
annotated dataset for Dari-language misinformation detection on
YouTube, and presented three principal findings.

\textbf{Structurally:} Information~Type and Harm~Level are not
independent. 55.9\% of Misinformation is associated with at least
Medium harm versus 1.0\% for True content, enabling Information~Type
classifiers to function as implicit harm-triage filters in content
moderation pipelines.

\textbf{Methodologically:} Pair-input encoding, which routes the
video title and description into separate BERT segments, yields a
+7.0~pp improvement in Misinformation recall---the safety-critical
minority class---over naive concatenation, despite modest overall
macro~F1 differences. This gain is practically meaningful in
harm-sensitive deployment contexts even when not statistically
significant at the 95\% level.

\textbf{Comparatively:} ParsBERT, a Dari/Farsi-specialised model,
shows consistent directional advantages over XLM-RoBERTa-base
across all metrics and class-level results, suggesting that
language-specialised pre-training confers meaningful benefits
for low-resource Dari classification.

Future work will pursue six directions: (1)~\emph{multimodal
integration}---incorporating audio speech recognition and visual
keyframe features to exploit cues unavailable in text metadata
alone; (2)~\emph{joint multi-task learning}---simultaneously
predicting Information~Type and Harm~Level to exploit their
structural coupling; (3)~\emph{external knowledge integration}---
connecting model predictions to fact-checking APIs and knowledge
graphs for claim-level verification; (4)~\emph{larger test sets}---
collecting sufficient held-out data to achieve statistical power
for the 2~pp differences observed; (5)~\emph{annotation refinement}---
sub-dividing the Partly~True category to reduce the boundary
ambiguity identified in the error analysis; and
(6)~\emph{cross-dialectal transfer}---leveraging larger Farsi
resources to improve Dari performance through targeted
domain adaptation.
We release DariMis-to the research community to support
progress along all of these directions.

\balance


\end{document}